\pdfoutput=1

\documentclass[conference,a4paper]{APSIPA2021}
\usepackage{amsmath}
\usepackage{graphicx}
\usepackage{multirow}
\usepackage{booktabs}
\usepackage{threeparttable}
\usepackage[style=numeric, backend=bibtex,style=ieee,]{biblatex}
\bibliography{mybib.bib}

\usepackage{geometry}
\usepackage{fancyhdr}

\fancypagestyle{firststyle}{
  \fancyhf{}
  \fancyhead[C]{2024 Asia Pacific Signal and Information Processing Association Annual Summit and Conference (APSIPA ASC)}
}

\begin{document}

\title{Analytic Study of Text-Free Speech Synthesis for Raw Audio using a Self-Supervised Learning Model}

% \author{
% \authorblockN{
% Thomas Chang\authorrefmark{1} and
% Yanning Zhang\authorrefmark{2}
% }

%\author{Joonyong Park$^1$, Daisuke Saito$^1$, Minematsu Nobuaki$^{1}$\\
\author{Joonyong Park$^*$, Daisuke Saito$^*$, Nobuaki Minematsu$^*$\\

\authorblockA{
\authorrefmark{1}
The University of Tokyo, Japan \\
E-mail: \{jpark, dsk\_saito, mine\}@gavo.t.u-tokyo.ac.jp}
}

\maketitle
\thispagestyle{firststyle}
\pagestyle{fancy}

\begin{abstract}
We examine the text-free speech representations of raw audio obtained from a self-supervised learning (SSL) model by analyzing the synthesized speech using the SSL representations instead of conventional text representations. Since raw audio does not have paired speech representations as transcribed texts do, obtaining speech representations from unpaired speech is crucial for augmenting available datasets for speech synthesis. Specifically, the proposed speech synthesis is conducted using discrete symbol representations from the SSL model in comparison with text representations, and analytical examinations of the synthesized speech have been carried out. The results empirically show that using text representations is advantageous for preserving semantic information, while using discrete symbol representations is superior for preserving acoustic content, including prosodic and intonational information.
\end{abstract}
\vspace{-1mm}

\section{Introduction}
\vspace{-1mm}

Current speech synthesis has significantly advanced through deep learning models, greatly surpassing the performance of traditional speech synthesis models~\cite{borgholt2022brief}. These models generally obtain speech feature vectors from the input text through an encoder, then output a Mel-spectrogram using methods such as Attention or Variational Inference, and finally convert it into speech through a Vocoder~\cite{shen2018natural,Ren2020FastSpeech2F,kim2021conditional}. At this stage, speech synthesis models essentially learn the correspondence between the training speech source and the 'input representation,' which describes the speech source and is traditionally represented as a transcribed text script~\cite{DBLP:conf/icassp/ShenPWSJYCZWRSA18}. 

However, a constraint of such models is that enhancing performance necessitates the additional task of pairing input representations as labels to the training speech source, which involves human effort and thus incurs significant costs. Furthermore, such text-based input representations vary by language, requiring even more resources when creating multilingual speech synthesis. To overcome these constraints, the use of Self-Supervised Learning (SSL) models, which can extract usable information from the raw speech source, is being considered. 

This study evaluates the performance of speech synthesis through several different input expressions, which analyzes the factors to consider when achieving such as "zero-resource synthetic speech", and evaluates the intelligibility, naturalness, and quality of the synthesized speech to see how each factor affects those speech.

\section{Prior Studies}
\vspace{-1mm}
\begin{figure}
    \centering
    \includegraphics[width=0.85\linewidth]{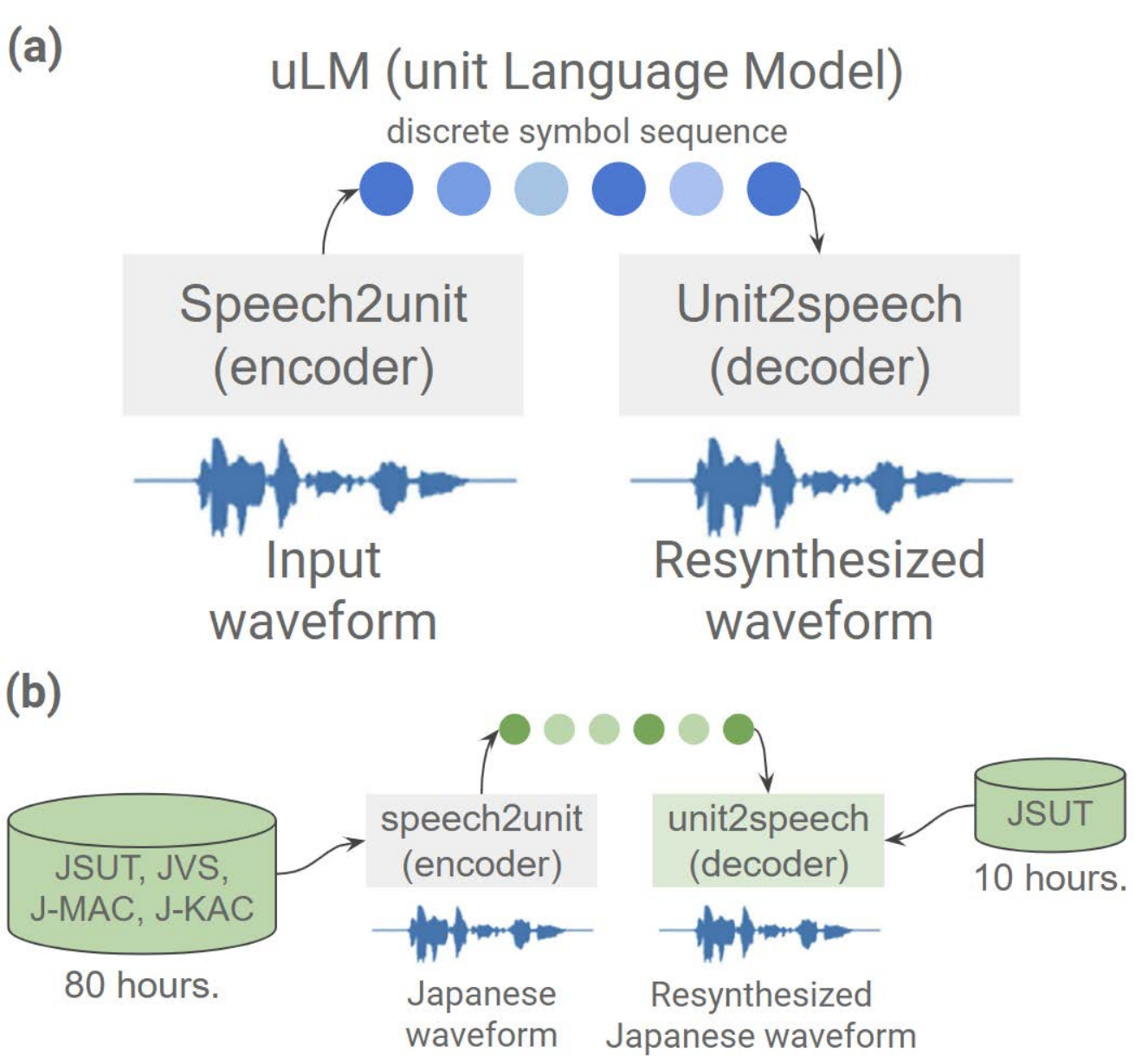}
    \caption{(a) Architecture of GSLM and (b) Application to Japanese Language}
    \vspace{-3mm}
\end{figure}

First, traditional speech synthesis models have focused on converting transcribed natural language text into a format more suitable for speech synthesis to obtain potential input representations. This includes text normalization, tokenization, the assignment of morphemes and parts of speech, phonemizing, and assigning accents to natural language texts through pre-prepared dictionaries and libraries. However, such methods have limitations in generalizing beyond single-language or high-resource language domains.

Therefore, methods that can obtain effective intermediate representations for speech synthesis tasks without text transcription, using audio data as a modality, have recently been researched. Self-supervised learning models are trained on a large amount of unlabeled audio data to predict hidden units or symbols not directly observed in the input audio data. Consequently, obtaining corresponding input representations for unlabeled speech has become more accurate, and it has also become possible to incorporate paralinguistic and non-verbal information into these input representations.

In a previous study on this approach, the Generative Spoken Language Modeling (GSLM)~\cite{lakhotia-etal-2021-generative} provided a methodology for speech synthesis from raw audio. Figure 1(a) shows a schematic illustration of the GSLM model architecture. GSLM processes and resynthesizes speech not through commonly used natural language texts but through discrete symbols, which are feature representations extracted from speech using SSL models. The model consists of an encoder called speech2unit, a decoder called unit2speech, and a unit language model (uLM). The speech2unit module converts audio waveforms into discrete symbols in text form through feature representations via SSL models like CPC~\cite{DBLP:journals/corr/abs-1807-03748}, wav2vec 2.0~\cite{NEURIPS2020_92d1e1eb}, and HuBERT~\cite{9585401}. It then quantizes the features with a predetermined codebook to obtain the discrete symbol sequence. The codebook is obtained by applying $k$-means clustering to the framewise features of the training data.

Conversely, the unit2speech module generates audio waveforms from sequences of discrete symbols using traditional text-to-speech models like Tacotron 2~\cite{DBLP:conf/icassp/ShenPWSJYCZWRSA18} and neural vocoder models. In this process, the speech synthesis model is trained with pairs of training speech and discrete symbols outputted by the encoder. In the case of speech resynthesis, there is a pipeline where the speech encoded through the speech2unit module is transformed back into synthesized speech through the unit2speech module.
However, since unit2speech is basically trained in a single language, language dependency needs to be resolved to use it for multiple languages. Figure 1(b) shows a Japanese application of the GSLM model architecture, where language dependence was resolved to some extent by training the speech2unit's $k$-means clustering model and unit2speech module with a Japanese dataset.
The uLM module, positioned between the two modules mentioned above, treats discrete symbols like character symbols and operates as a language model using the Transformer~\cite{NIPS2017_3f5ee243} network.

Additionally, as a prior study on the evaluation of synthesized speech without labels, the Zero Resource Speech Challenge (ZRC) can be mentioned. The research sets several tasks with the goal of constructing language processing models using only audio data by removing text labels from a ground truth speech corpus. The objective is to surpass the topline models, which are trained from transcribed text, in metrics for each task. Among the tasks, in the 'Discrete Resynthesis' task, the input representations obtained through SSL models are re-synthesized, and clarity and naturalness are evaluated through Character Error Rate (CER) and Mean Opinion Score (MOS). The experimental results have shown that models adopting certain SSL methods achieved better MOS results than the topline, and it was also confirmed that the more bit-information the input representations contained, the higher the metrics of synthesized speech were.~\cite{dunbar2020zero,dunbar2021zero,Dunbar_2022}

In addition, through structures such as GSLM, research on solving audio tasks as end-to-end using untranscribed audio data has continued to show superior results compared to previous baselines~\cite{Niekerk2021ACO,10096797,kim23k_interspeech}.
From such prior studies, it is suggested that by obtaining input representations through self-supervised learning methods for audio data without pre-transcribed labels, it is possible that SSL representations may show superior performance compared to text transcriptions for speech synthesis in general. 

\section{Comparative Study on Methods Based on Input Representations}
\vspace{-1mm}

In the case of the Zero Resource Speech Challenge, only the evaluation of intelligibility and naturalness after synthesis was present, while the quality of the acoustics was not evaluated. Additionally, prior studies have not conducted complete analysis of multiple languages or multiple encoder-decoder pairs, and this has not been analyzed with several conditions such as symbol size and output layer of Transformer.

Therefore, this study evaluates by adding metrics that can measure acoustic quality, in addition to the experimental results on clarity and naturalness, along with changes in the models used. Furthermore, SSL models are often trained in a specific language, which could show degraded performance in speech synthesis for multiple languages due to language dependency. Hence, this research proceeds with experiments not only in English but also in Japanese, investigates the presence or absence of language dependency in the representations outputted through SSL models, and ultimately analyzes the impact of such factors on the nature of synthesized speech.

\subsection{Experiment Setting}
\vspace{-1mm}

\begin{figure}[tb]
    \centering
    \includegraphics[width=0.95\linewidth]{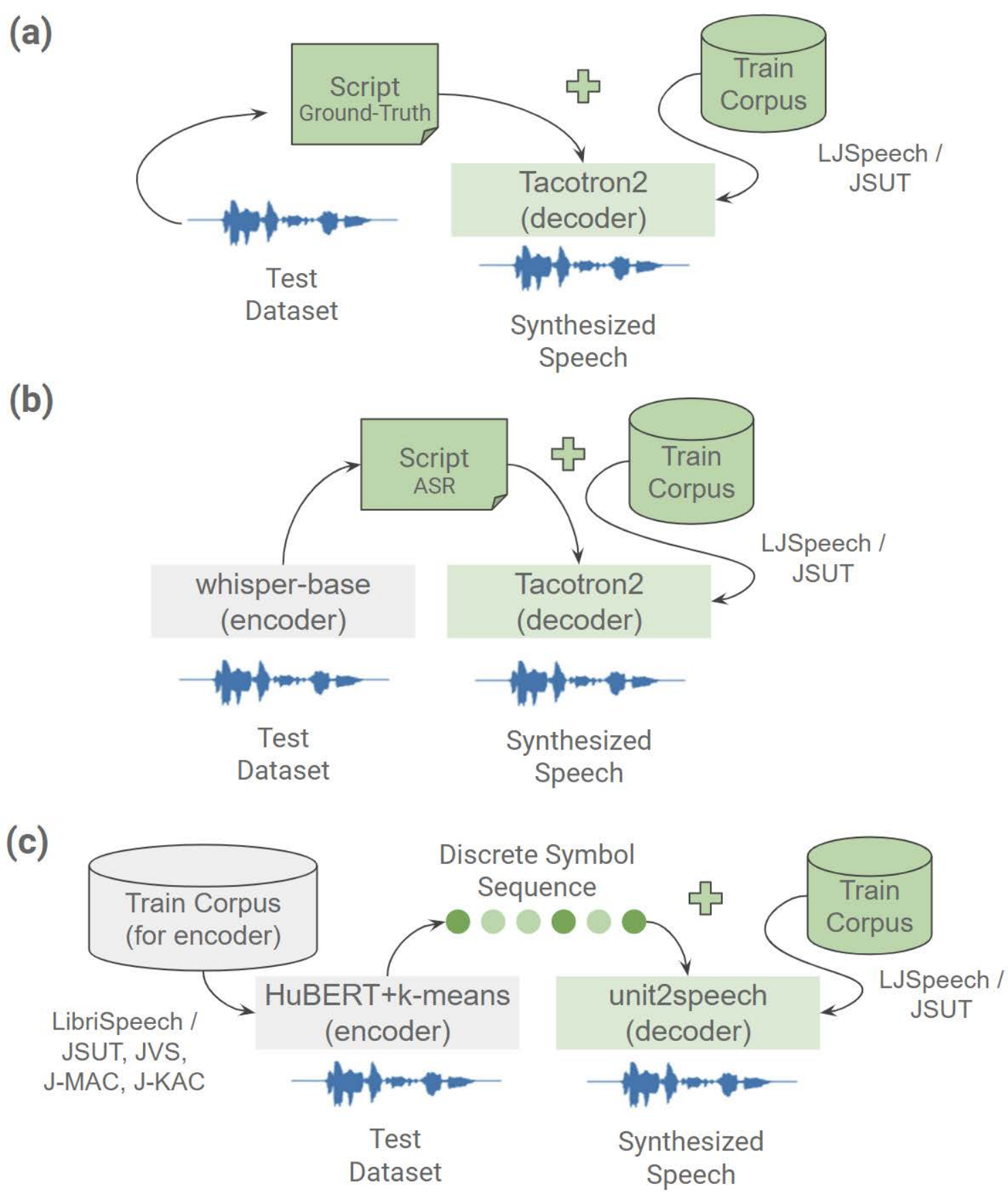}
    \caption{Building a speech synthesis system using (a) ground-truth script labels, (b) speech recognition (ASR) model, and (c) self-supervised learning (SSL) model}
    \label{fig:fig/ronbun_gslm_config.pdf}
\end{figure}

In this experiment, models for three different input representations were created for two languages: English and Japanese. The structure is shown in Figure 2.

\begin{table}[tb]
  \centering
  \caption{Corpora used for training hypothesis model}
  \label{tab:corpus}
  \resizebox{\columnwidth}{!}{
    \begin{tabular}{cccc}
      \toprule
      Language                 & S2u(multiple speakers)                                                                               & u2S(single speaker)     \\\midrule % & (S2u/u2S)                \\ \hline
      English                  & LibriSpeech\cite{7178964}                                                                              & LJSpeech\cite{ljspeech} \\        %& 100h/11h                 \\
      Japanese                 & \begin{tabular}{@{}c@{}}Reazonspeech~\cite{reasonspeech} (HuBERT) \\ JSUT\cite{DBLP:journals/corr/abs-1711-00354}, JVS\cite{DBLP:journals/corr/abs-1908-06248}, JKAC\cite{jkac}, JMAC\cite{DBLP:conf/interspeech/TakamichiNTS22} ($k$-means) \end{tabular}   & JSUT                \\      %& 75h/10h                  \\
      \bottomrule
    \end{tabular}
  }
\end{table}

\textbf{The first model} (2-(a)) is used as a reference, synthesizing speech using text scripts as input representations, which serve as the correct (gold) labels.

\textbf{The second model} (2-(b)) serves as a baseline, synthesizing speech after recognizing the training speech dataset through an ASR model. The ASR model used is Whisper~\cite{radford2022whisper}, with the base model used for both English and Japanese. For these above two models, preprocessing includes the use of the English cleaner provided by Tacotron2 for English, and morphological analysis through Mecab for Japanese.

\textbf{The third model} (2-(c)) is used as a hypothesis, synthesizing speech by obtaining input representations in discrete symbol form via the speech2unit module from the self-supervised learning model presented by GSLM. \footnote{As this experiment focuses only on speech resynthesis, which does not require the uLM, we only analyze the speech2unit and unit2speech equivalent modules.} For this, pipelines for English and Japanese are presented. The datasets for each language used in speech2unit and unit2speech are shown in Table 1; the sampling rate of the audio was unified to 16k for all pre-trained, training, and test datasets.
We use HuBERT-base for outputting the input representation in the form of a discrete symbol through a $k$-means model. 

In addition, several differential elements were introduced in the SSL model and the model was analyzed using the ablation method. Firstly, in order to analyze the language dependency of speech2unit, the difference between the two synthetic speech is analyzed as shown in Figure 3, assuming that speech2unit is trained in the same language as unit2speech and in a different language. This creates four combinations of S2u-u2S pair in total.
Secondly, in order to evaluate the effect of changing the code length of the discrete symbol, synthetic speech that is synthesized from three different discrete symbols are compared, by training the $k$-means model's clustering number to 50, 200, and 1000, respectively. For all cases, processing was performed to remove repeated symbols from the discrete symbol sequences if length $n$ or more is continuous for each symbol. For this experiment, $n$ was fixed to 1.
Thirdly, to identify differences in output between Transformer layers within speech2unit, synthetic speech from the representation by speech2unit's different layers is been compared: in this experiment, 6th and 12th layer. Attempts have been conducted to find effective features within SSL models, and it has also been found that the differences between these layers has made difference in 'linguistic' and 'acoustic' factors~\cite{9746250,qian22b}. These conditions also verify that the prior research is consistent with the variables of language and code length.

In common with all three models, the Tacotron 2 model was used for generating synthesized speech through input representations. The model is designed to train Mel-Spectrograms from the text of input representations and to output speech using a Vocoder under the same conditions for each language, except for the differences in input representations. 

\begin{figure}[tb]
    \centering
    \includegraphics[width=0.85\linewidth]{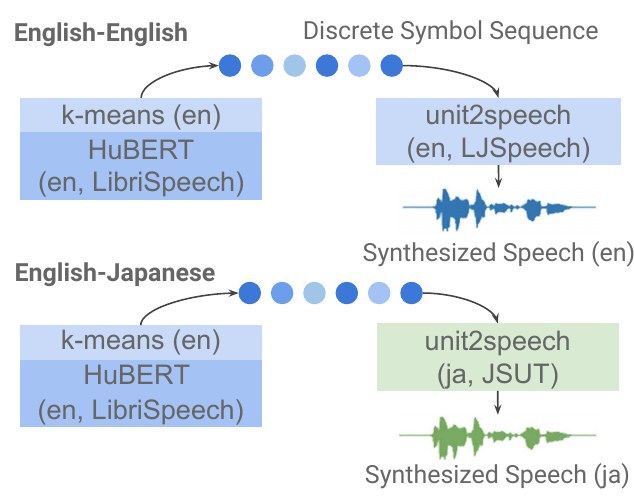}
    \caption{An example of a SSL system that speech2unit-unit2speech pairs are language-matched and unmatched}
    \label{fig:fig/ronbun_gslm_config2.pdf}
    \vspace{-3mm}
\end{figure}

For the test datasets, 100 utterances selected from LibriSpeech-dev were used for English, and for Japanese, 100 utterances selected from utterances not used as training data in the JSSS corpus~\cite{DBLP:journals/corr/abs-2010-01793} were used. 
The test dataset is also converted into the form of same representations that are used in the training phase, such as transcribed text (GT), ASR script (ASR), and discrete symbol sequence (SSL) respectively. Those input representations are used for inference.

\section{Experiment Results}
\vspace{-1mm}

The experiment is broadly divided into evaluations of speech language and acoustic quality. The speech language evaluation aims to assess changes in the linguistic semantic elements of synthesized speech based on input representations, while the acoustic quality evaluation focuses on elements other than the language contained in the synthesized speech audio.

The data in Table 2 are the result of SSL speech synthesis comparing with the reference GT and baseline ASR. The characters in the item represents 'S2u-u2S language matching', 'symbol size', and 'Transformer output layer', respectively. For example, in the case of \emph{match-200-L12}, it refers a synthesized speech that used same S2u with u2S, 200 symbol size and discrete symbols obtained from the 12th layer.

\begin{table*}[t]
    \centering
    \label{}
    \caption{Metrics comparing Synthesized Speech from the input representation of Ground Truth, ASR Model, and various SSL models for English and Japanese respectively. 
    Item is emboldened when it recorded the most superior value among SSL models, and baseline is underlined if the baseline value is superior than the item.}
    \resizebox{\linewidth}{!}
    {
        \begin{tabular}{lccccc|cccccc}
            \toprule
            & \multicolumn{5}{c}{English} & \multicolumn{5}{c}{Japanese} \\
            & {WER(\%)$\downarrow$} & {PER(\%)$\downarrow$} &{UTMOS$\uparrow$} &{WARP-Q$\uparrow$} &{SDR(dB)$\uparrow$} & {CER(\%)$\downarrow$} & {PER(\%)$\downarrow$} &{UTMOS$\uparrow$} &{WARP-Q$\uparrow$} &{SDR(dB)$\uparrow$} \\
            \midrule
            {GT} & 4.22 & 8.14 & 3.93 ± 0.067 & - & - & 20.68 & 41.59 & 2.68 ± 0.045 & - & - \\ 
            {ASR} & \underline{5.41} & \underline{9.92} & 3.41 ± 0.084 & 2.71 & -18.55 & \underline{24.62} & \underline{44.91} & 2.57 ± 0.049 & 2.23 & -23.53 \\ 
            
            \midrule
            {match-50-L6} & 8.37 & 19.09 & 3.31 ± 0.096 & 2.62 & -19.91 & 42.88 & 82.19 & 2.10 ± 0.064 & 2.20 & -23.95  \\
            {match-50-L12} & 7.82 & 16.06 & 3.49 ± 0.076 & 2.78 & -19.32 & 35.73 & 69.02 & 2.21 ± 0.055 & 2.29 & -23.81 \\ 
            {match-200-L6} & 6.95 & 15.24 & 3.45 ± 0.068 & 2.86 & -17.14 & 32.95 & 59.08 & 2.44 ± 0.041 & 2.37 & -23.56 \\ 
            {match-200-L12}& 6.63 & 14.53 & 3.60 ± 0.057 & 2.89 & -17.05 & 27.04 & 50.07 & 2.62 ± 0.053 & 2.35 & -23.41 \\ 
            {match-1000-L6}& 6.25 & 13.09 & 3.63 ± 0.060 & \textbf{2.93} & -17.74 & 29.54 & 53.35 & 2.45 ± 0.051 & 2.36 & -23.45 \\ 
            {match-1000-L12}& \textbf{5.45} & \textbf{10.37} & \textbf{3.72 ± 0.063}  & 2.92 & -17.37 & \textbf{26.58} & \textbf{49.40} & \textbf{2.64 ± 0.043}  & 2.35 & -23.39 \\ 
            {unmatch-50-L6}& 10.25 & 25.42 & 3.29 ± 0.093 & 2.74 & -19.44 & 43.67 & 83.75 & 2.12 ± 0.067 & 2.18 & -23.92 \\ 
            {unmatch-50-L12}& 8.90 & 20.51 & 3.41 ± 0.079 & 2.69 & -18.94 & 38.06 & 77.29 & 2.18 ± 0.059 & 2.21 & -23.61 \\ 
            {unmatch-200-L6}& 8.84 & 20.13 & 3.37 ± 0.069 & 2.78 & -17.46 & 33.58 & 62.76 & 2.43 ± 0.053 & 2.33 & -23.80 \\ 
            {unmatch-200-L12}& 7.37 & 15.42 & 3.47 ± 0.058 & 2.84 & \textbf{-16.73} & 28.22 & 52.32 & 2.59 ± 0.047 & 2.37 & -23.44 \\ 
            {unmatch-1000-L6}& 6.31 & 13.96 & 3.61 ± 0.057& 2.80 & -17.09 & 31.83 & 59.28 & 2.45 ± 0.054 & 2.36 & \textbf{-23.07} \\ 
            {unmatch-1000-L12}& 6.29 & 13.59 & 3.64 ± 0.065 & 2.87 & -16.69 & 27.45 & 50.81 & 2.57 ± 0.048 & \textbf{2.40} & -23.47 \\ 

            \bottomrule
        \end{tabular}
    }
\end{table*}

\subsection{Speech Intelligibility}
\vspace{-1mm}

For the language evaluation, the impact of each input representation on the intelligibility of speech language was examined by investigating the error rates for the target languages. The output from Whisper-base and the correct scripts of each test dataset were compared, investigating Word Error Rate (WER) for English and Character Error Rate (CER) for Japanese. As an initial step in evaluating the performance of the speech recognition model used in the experiment, the error rate was investigated against the correct scripts for the GT test dataset processed by speech recognition. The results showed a word error rate of 2.41~\% for English and a character error rate of 19.35~\% for Japanese.

Also, to perform a more detailed analysis using a language-independent metric, phonemes were obtained from each piece of speech information, and their errors were also investigated. The Phoneme Error Rate (PER) was investigated using the speech of the GT test dataset as a reference, with each speech is analyzed as hypothesis. We used the phoneme recognizer, \emph{allosaurus}~\cite{9054362}, to obtain phoneme directly from the speech.

The average error rates obtained for speech synthesized from ground-truth scripts, ASR model, and SSL model input representations are shown in Table 2. For both languages, speech synthesis using the GT script showed the lowest error rate. On the other hand, for label-less cases, synthesis through the ASR model showed lower error rates than synthesis through the SSL model. 
Also, it was confirmed that the relationship between PER and WER or CER showed a proportional relationship with all data. This shows that the intelligibility consistently affects the errors from low-level elements like phonemes to high-level elements like words. These results suggest that using input representations obtained through ASR, which is natural text, leads to better conveyance of linguistic elements like intelligibility. This aligns with the tendency in the ZRC's prior study, where the topline using text transcription for both English and other languages showed superior ABX values in the ABX task used for phoneme discrimination~\cite{Dunbar_2022}.

Moreover, regarding the intelligibility of synthesized speech through language matching, the language-matched model's results showed slightly lower error rates than the unmatched model's, showing that language dependence of SSL model has non-negligible consequences.
For the codec length, performance improved as the codec length of $k$-means clustering increased, reaching a level comparable to the ASR result in the case of English WER. In addition, in all cases, the synthesized speech through the 12th layer was able to obtain a lower error rate than the synthesized speech through the 6th layer. This is consistent with the results of previous studies that the layer close to the final output has more 'semantic' information than the intermediate layer.~\cite{9746250}

\subsection{Speech Naturalness}
\vspace{-1mm}

In the case of ZRC, the subjective metric Mean Opinion Score (MOS) was used to evaluate naturalness, which can be considered an indicator that comprehensively evaluates both linguistic and extra-linguistic information. On the other hand, this study employed the pre-trained UTMOS model. UTMOS, trained on multiple languages including English, predicts automated MOS. UTMOS is evaluated independently without a comparative subject, allowing for the analysis of absolute linguistic and paralinguistic information, but making it impossible to compare voice changes between GT and synthesized speech.

The average UTMOS values obtained for speech synthesized from correct labels, ASR model, and SSL model input representations are shown in Table 2. \footnote{Although 95~\% confidence level was indicated in the table, as UTMOS is not a subjective value but an objective value from the model, the value was calculated from the standard deviation of the sample.} Common across all languages, speech synthesis using correct scripts showed the highest values. Meanwhile, under label-less synthesis conditions, the MOS for synthesized speech from SSL models was slightly higher across all language conditions than for those from ASR models, with this difference being greater in English synthesized speech than in Japanese. Additionally, speech2unit-unit2speech pairs of the same language showed predominant MOS scores for both languages.\footnote{The difference in absolute UTMOS values between Japanese and English can be attributed to the UTMOS evaluation model being primarily trained on English speech data.} 

Moreover, the performance improved as codec length increased; however, while there was an big increase without exception when the token increasing from 50 to 200, there was a mild increase or even decrease when the token increasing from 200 to 1000. Thus, it can be said that naturalness is correlated to intelligibility, but this correlation decreases as the number of tokens increases. Also, the output obtained in the 12th layer obtained better results in terms of naturalness compared to those obtained in the 6th layer, following the trend has shown in the intelligibility: while some errors are within confidence levels.

These results propose a hypothesis that SSL discrete representations containing paralinguistic information like accents and intonations can enhance the naturalness of synthesized speech more than input representations from ASR models, suggesting that this can change depending on the language dependency of the SSL model and the amount of target language data it was trained on. It also suggests that increasing the language dependency of SSL models could incorporate more language-appropriate paralinguistic information within discrete representations. This hypothesis aligns with the results in ZRC, where the synthesized speech results for Indonesian, a language not trained in ZRC, more frequently failed to surpass the MOS of top-line models utilizing text transcription for English, a trained language~\cite{Dunbar_2022}.

\subsection{Audio Quality and Noisiness}
\vspace{-1mm}

Following the evaluation of linguistic and naturalness aspects, which are related to speech language, the overall acoustic elements of the synthesized speech were assessed through quality evaluation.

Initially, the quality of digital audio was evaluated based on codecs, considering input representations as compressed audio representations, which can be viewed as neural speech codecs. Therefore, assuming the speech synthesis system as a single system, the output audio was analyzed as degraded audio compared to a reference. For such evaluations, PESQ is a representative metric, but in this case, speaker information and speech type can influence the results. Thus, to compensate for temporal mismatches between GT and resynthesized signals internally and to be resilient against errors commonly occurring in audio codecs, the WARP-Q metric was added to the evaluation. To address this, synthesized speech using correct scripts was used as a reference, and synthesized speech using ASR and SSL was compared as hypotheses.

The average values of WARP-Q~\cite{Wissam_IET_Signal_Process2022} obtained for speech synthesized from ASR model and SSL model input representations are shown in Table 2. Generally, synthesized speech generated through the SSL model showed higher metrics compared to that synthesized through the ASR model. However, although there was a slight tendency due to Transformer layer and token length, differences due to the SSL model's encoder did not show a consistent trend across languages or tasks, with only minor increases or decreases observed.

The quality of audio in terms of noise inclusion was also evaluated. Signal Distortion Rate (SDR), although an indicator used in source separation, can be considered for evaluating the degree of noise inclusion in the output audio relative to the input audio, hence indicative of acoustic quality. 
The average SDR values obtained for speech synthesized from ASR model and SSL model input representations are shown in the table. Overall, although the values were low, synthesized speech through the SSL model showed slightly better metrics. Moreover, regarding differences due to the SSL model's encoder, systems crossing languages for English synthesized speech showed better results, while for Japanese synthesized speech, systems not crossing languages did better, indicating no clear trend. There is little trend for Transformer layer and token length as well.

Thus, speech synthesis models generated using speech recognition showed slightly more codec distortion and noise compared to those using SSL models, and synthesized speech's dependency on token language did not bring significant differences in speech quality. 

\section{Conclusion}
\vspace{-1mm}

This study investigated input representations in speech synthesis systems and created synthesized speech through systems constructed using each type of input representation. These were then compared and analyzed across languages and representations through newly considered metrics and results from prior research.

The findings revealed that no input representation demonstrated higher metrics than correct labels. Natural language input representations through speech recognition models showed dominance in linguistic vocal aspects, while discrete symbol input representations through self-supervised learning models were superior in aspects of naturalness and acoustic quality. Furthermore, the study highlighted differences in self-supervised learning model performance based on language dependency initiated by the language of the data. 
Also, the metrics shown overall improvement as the codec length of $k$-means clustering increased, more dramatically in the speech language metrics than the acoustic metrics. Furthermore, regardless of the language using, it is able to be found that Transformer layers close to the final output in linguistic elements had more semantic information than those that did not, and it was possible to demonstrate the tendency of previous studies.

Future efforts will focus on devising methods to minimize such language dependency. The goal is to advance the evaluation of tasks in more multilingual contexts than currently possible by transforming text into language-informed discrete tokens through tokenizers without preprocessing the text according to language-specific rules.

\printbibliography

@article{borgholt2022brief,
  author       = {Lasse Borgholt and
                  Jakob Drachmann Havtorn and
                  Joakim Edin and
                  Lars Maal{\o}e and
                  Christian Igel},
  title        = {A Brief Overview of Unsupervised Neural Speech Representation Learning},
  journal      = {CoRR},
  volume       = {abs/2203.01829},
  year         = {2022},
  eprinttype    = {arXiv},
  eprint       = {2203.01829},
}

@misc{shen2018natural,
      title={Natural TTS Synthesis by Conditioning WaveNet on Mel Spectrogram Predictions}, 
      author={Jonathan Shen and Ruoming Pang and Ron J. Weiss and Mike Schuster and Navdeep Jaitly and Zongheng Yang and Zhifeng Chen and Yu Zhang and Yuxuan Wang and RJ Skerry-Ryan and Rif A. Saurous and Yannis Agiomyrgiannakis and Yonghui Wu},
      year={2018},
      eprint={1712.05884},
      archivePrefix={arXiv}
}

@inproceedings{kim2021conditional,
  author       = {Jaehyeon Kim and
                  Jungil Kong and
                  Juhee Son},
  title        = {Conditional Variational Autoencoder with Adversarial Learning for
                  End-to-End Text-to-Speech},
  booktitle    = {Proceedings of the 38th International Conference on Machine Learning
                  },
  volume       = {139},
  pages        = {5530--5540},
  year         = {2021},
}

@inproceedings{NIPS2017_3f5ee243,
 author = {Vaswani, Ashish and Shazeer, Noam and Parmar, Niki and Uszkoreit, Jakob and Jones, Llion and Gomez, Aidan N and Kaiser, \L ukasz and Polosukhin, Illia},
 booktitle = {Advances in Neural Information Processing Systems},
 publisher = {Curran Associates, Inc.},
 title = {Attention is All you Need},
 volume = {30},
 year = {2017}
}

@inproceedings{dunbar2020zero,
  author={Ewan Dunbar and Julien Karadayi and Mathieu Bernard and Xuan-Nga Cao and Robin Algayres and Lucas Ondel and Laurent Besacier and Sakriani Sakti and Emmanuel Dupoux},
  title={{The Zero Resource Speech Challenge 2020: Discovering Discrete Subword and Word Units}},
  year=2020,
  booktitle={Proc. Interspeech 2020},
  pages={4831--4835},
}

@inproceedings{dunbar2021zero,
      title={The Zero Resource Speech Challenge 2021: Spoken language modelling}, 
      author={Ewan Dunbar and Mathieu Bernard and Nicolas Hamilakis and Tu Anh Nguyen and Maureen de Seyssel and Patricia Rozé and Morgane Rivière and Eugene Kharitonov and Emmanuel Dupoux},
      year={2021},
      booktitle={Proc. Interspeech 2021},
      pages={1574--1578},
}

@article{Dunbar_2022,
  
	year = 2022,
  
	publisher = {Institute of Electrical and Electronics Engineers ({IEEE})},
  
	volume = {16},
  
	number = {6},
  
	pages = {1211--1226},
  
	author = {Ewan Dunbar and Nicolas Hamilakis and Emmanuel Dupoux},
  
	title = {Self-Supervised Language Learning From Raw Audio: Lessons From the Zero Resource Speech Challenge},
  
	journal = {{IEEE} Journal of Selected Topics in Signal Processing}
}

@article{lakhotia-etal-2021-generative,
    title = "On Generative Spoken Language Modeling from Raw Audio",
    author = "Lakhotia, Kushal  and
      Kharitonov, Eugene  and
      Hsu, Wei-Ning  and
      Adi, Yossi  and
      Polyak, Adam  and
      Bolte, Benjamin  and
      Nguyen, Tu-Anh  and
      Copet, Jade  and
      Baevski, Alexei  and
      Mohamed, Abdelrahman  and
      Dupoux, Emmanuel",
    journal = "Transactions of the Association for Computational Linguistics",
    volume = "9",
    year = "2021",
    address = "Cambridge, MA",
    publisher = "MIT Press",
    pages = "1336--1354",
}

@article{DBLP:journals/corr/abs-1807-03748,
  author    = {A{\"{a}}ron van den Oord and
               Yazhe Li and
               Oriol Vinyals},
  title     = {Representation Learning with Contrastive Predictive Coding},
  journal   = {CoRR},
  volume    = {abs/1807.03748},
  year      = {2018},
  eprinttype = {arXiv},
  eprint    = {1807.03748},
}

@inproceedings{NEURIPS2020_92d1e1eb,
 author = {Baevski, Alexei and Zhou, Yuhao and Mohamed, Abdelrahman and Auli, Michael},
 booktitle = {Advances in Neural Information Processing Systems},
 title = {wav2vec 2.0: A Framework for Self-Supervised Learning of Speech Representations},
 volume = {33},
 year = {2020},
 pages = {12449--12460}
}

@ARTICLE{9585401,
  author={Hsu, Wei-Ning and Bolte, Benjamin and Tsai, Yao-Hung Hubert and Lakhotia, Kushal and Salakhutdinov, Ruslan and Mohamed, Abdelrahman},
  journal={IEEE/ACM Transactions on Audio, Speech, and Language Processing}, 
  title={Hu{BERT}: Self-Supervised Speech Representation Learning by Masked Prediction of Hidden Units}, 
  year={2021},
  volume={29},
  number={},
  pages={3451-3460}}

@INPROCEEDINGS{7178964,
  author={Panayotov, Vassil and Chen, Guoguo and Povey, Daniel and Khudanpur, Sanjeev},
  booktitle={2015 IEEE International Conference on Acoustics, Speech and Signal Processing}, 
  title={Libri{S}peech: An {ASR} corpus based on public domain audio books}, 
  year={2015},
  volume={},
  number={},
  pages={5206-5210}
  }

@misc{radford2022whisper,
      title={Robust Speech Recognition via Large-Scale Weak Supervision.}, 
      author={Radford, Alec and Jong Wook, Kim and Tao, Xu and Greg, Brockman and Christine, McLeavey and Ilya, Sutskever},
      howpublished = {\url{https://cdn.openai.com/papers/whisper.pdf}},
      year={2022},
}

@article{Wissam_IET_Signal_Process2022,
  author = {Jassim, Wissam A. and Skoglund, Jan and Chinen, Michael and Hines, Andrew},
  title = {Speech quality assessment with {WARP-Q}: From similarity to subsequence dynamic time warp cost},
  journal = {IET Signal Processing},
  volume = {16},
  number = {9},
  pages = {1050–-1070},
  year = {2022},
 }

@article{DBLP:journals/corr/abs-1711-00354,
  author    = {Ryosuke Sonobe and
               Shinnosuke Takamichi and
               Hiroshi Saruwatari},
  title     = {{JSUT} corpus: free large-scale {J}apanese speech corpus for end-to-end
               speech synthesis},
  journal   = {CoRR},
  volume    = {abs/1711.00354},
  year      = {2017},
  eprinttype = {arXiv},
  eprint    = {1711.00354},
}

@article{DBLP:journals/corr/abs-1908-06248,
  author    = {Shinnosuke Takamichi and
               Kentaro Mitsui and
               Yuki Saito and
               Tomoki Koriyama and
               Naoko Tanji and
               Hiroshi Saruwatari},
  title     = {{JVS} corpus: free {J}apanese multi-speaker voice corpus},
  journal   = {CoRR},
  volume    = {abs/1908.06248},
  year      = {2019},
  eprinttype = {arXiv},
  eprint    = {1908.06248},
}

@inproceedings{DBLP:conf/interspeech/TakamichiNTS22,
  author    = {Takamichi, Shinnosuke and
               Wataru, Nakata and
               Naoko, Tanji and
               Hiroshi, Saruwatari},
  title     = {{J-MAC:} {J}apanese multi-speaker audiobook corpus for speech synthesis},
  booktitle = {Interspeech 2022, 23rd Annual Conference of the International Speech
               Communication Association},
  pages     = {2358--2362},
  publisher = {{ISCA}},
  year      = {2022},
}

@inproceedings{jkac,
      author={Wataru Nakata and
              Tomoki Koriyama and
              Shinnosuke Takamichi and 
              Naoko Tanji and
              Yusuke Ijima and
              Ryo Masumura and 
              Hiroshi Saruwatari},
      title={Audiobook Speech Synthesis Conditioned by Cross-Sentence Context-Aware Word Embeddings}, 
      booktitle = {{Proc. The 11th ISCA SSW}},
      year={2021}
}

@inproceedings{DBLP:conf/icassp/ShenPWSJYCZWRSA18,
  author    = {Jonathan Shen and
               Ruoming Pang and
               Ron J. Weiss and
               Mike Schuster and
               Navdeep Jaitly and
               Zongheng Yang and
               Zhifeng Chen and
               Yu Zhang and
               Yuxuan Wang and
               RJ{-}Skerrv Ryan and
               Rif A. Saurous and
               Yannis Agiomyrgiannakis and
               Yonghui Wu},
  title     = {Natural {TTS} Synthesis by Conditioning {W}avenet on MEL-Spectrogram
               Predictions},
  booktitle = {2018 {IEEE} International Conference on Acoustics, Speech and Signal
               Processing},
  pages     = {4779--4783},
  year      = {2018},
}

@article{DBLP:journals/corr/abs-2010-01793,
  author       = {Shinnosuke Takamichi and
                  Mamoru Komachi and
                  Naoko Tanji and
                  Hiroshi Saruwatari},
  title        = {{JSSS:} free Japanese speech corpus for summarization and simplification},
  journal      = {CoRR},
  volume       = {abs/2010.01793},
  year         = {2020},
  eprinttype    = {arXiv},
  eprint       = {2010.01793},
}

@misc{reasonspeech,
  author    = {Yue, Yin and Daijiro, Mori and Seiji, Fujimoto},
  title     = {ReazonSpeech: A free and massive corpus for {J}apanese {ASR}},
  journal   = {Proceedings of Annual Meeting of the Association for NLP},
  howpublished = {\url{https://research.reazon.jp/_static/reazonspeech_nlp2023.pdf}},

  year      = {2023},
  pages    = {1134–1139},
}

@misc{ljspeech,
	author = {Ito, Keith and Johnson, Linda},
	howpublished = {\url{https://keithito.com/LJ-Speech-Dataset/}},
	title = {The LJ Speech Dataset},
	year = {2017}}

@article{Ren2020FastSpeech2F,
  title={FastSpeech 2: Fast and High-Quality End-to-End Text to Speech},
  author={Yi Ren and Chenxu Hu and Xu Tan and Tao Qin and Sheng Zhao and Zhou Zhao and Tie-Yan Liu},
  journal={ArXiv},
  year={2020},
  volume={abs/2006.04558},
}

@INPROCEEDINGS{9054362,
  author={Li, Xinjian and Dalmia, Siddharth and Li, Juncheng and Lee, Matthew and Littell, Patrick and Yao, Jiali and Anastasopoulos, Antonios and Mortensen, David R. and Neubig, Graham and Black, Alan W and Metze, Florian},
  booktitle={ICASSP 2020 - 2020 IEEE International Conference on Acoustics, Speech and Signal Processing (ICASSP)}, 
  title={Universal Phone Recognition with a Multilingual Allophone System}, 
  year={2020},
  volume={},
  number={},
  pages={8249-8253}}

@inproceedings{9746250,
  author={Kumar, Pratik and Sukhadia, Vrunda N. and Umesh, S.},
  booktitle={ICASSP 2022 - 2022 IEEE International Conference on Acoustics, Speech and Signal Processing (ICASSP)}, 
  title={Investigation of Robustness of Hubert Features from Different Layers to Domain, Accent and Language Variations}, 
  year={2022},
  volume={},
  number={},
  pages={6887-6891}}

@inproceedings{qian22b,
  title = 	 {{C}ontent{V}ec: An Improved Self-Supervised Speech Representation by Disentangling Speakers},
  author =       {Qian, Kaizhi and Zhang, Yang and Gao, Heting and Ni, Junrui and Lai, Cheng-I and Cox, David and Hasegawa-Johnson, Mark and Chang, Shiyu},
  booktitle = 	 {Proceedings of the 39th International Conference on Machine Learning},
  pages = 	 {18003--18017},
  year = 	 {2022}}

@article{Niekerk2021ACO,
  title={A Comparison of Discrete and Soft Speech Units for Improved Voice Conversion},
  author={Benjamin van Niekerk and Marc-Andr{\'e} Carbonneau and Julian Za{\"i}di and Matthew Baas and Hugo Seut{\'e} and Herman Kamper},
  journal={ICASSP 2022 - 2022 IEEE International Conference on Acoustics, Speech and Signal Processing (ICASSP)},
  year={2021},
  pages={6562-6566}}

@INPROCEEDINGS{10096797,
  author={Li, Xinjian and Jia, Ye and Chiu, Chung-Cheng},
  booktitle={ICASSP 2023 - 2023 IEEE International Conference on Acoustics, Speech and Signal Processing (ICASSP)}, 
  title={Textless Direct Speech-to-Speech Translation with Discrete Speech Representation}, 
  year={2023},
  volume={},
  number={},
  pages={1-5},
  doi={10.1109/ICASSP49357.2023.10096797}}

@inproceedings{kim23k_interspeech,
  author={Heeseung Kim and Sungwon Kim and Jiheum Yeom and Sungroh Yoon},
  title={{UnitSpeech: Speaker-adaptive Speech Synthesis with Untranscribed Data}},
  year=2023,
  booktitle={Proc. INTERSPEECH 2023},
  pages={3038--3042},
  doi={10.21437/Interspeech.2023-2326},
  issn={2958-1796}
}

\end{document}